\documentclass[lettersize,journal]{IEEEtran}
\usepackage{amsmath,amsfonts}
\usepackage{algorithmic}
\usepackage{algorithm}
\usepackage{array}
\usepackage[caption=false,font=normalsize,labelfont=sf,textfont=sf]{subfig}
\usepackage{textcomp}
\usepackage{stfloats}
\usepackage{url}
\usepackage{verbatim}
\usepackage{graphicx}
\usepackage{cite}
\usepackage{xcolor}
\usepackage{subfig}
\usepackage{multirow}
\usepackage[most]{tcolorbox}
\usepackage{hyperref}
\usepackage{booktabs}

\begin{document}

\title{E-SMAD: A Simple yet Effective Ensemble of Student Models \\ for Adversarial Defense}

\author{Muhammad Luqman Naseem,
        Yamin Hu, Yujiang Liu, Zhijian Chen
        and~Wenjian Luo,~\IEEEmembership{Senior Member,~IEEE}
\thanks{Muhammad Luqman Naseem, Yamin Hu, Yujiang Liu, Zhijian Chen, and Wenjian Luo are with Guangdong Provincial Key Laboratory of Novel Security Intelligence Technologies, Institute of Cyberspace Security, School of Computer Science and Technology, Harbin Institute of Technology, Shenzhen 518055, China. (e-mail: 22bf51040@stu.hit.edu.cn; huyamin@hit.edu.cn; 23S151125@stu.hit.edu.cn; 21B951010@stu.hit.edu.cn; luowenjian@hit.edu.cn.)}
\thanks{This study is supported by the National Natural Science Foundation of China (Grant No. 62576121), University Stability Support Program of Shenzhen (Grant No. GXWD20231130113127003), Shenzhen Science and Technology Program (Grant No. ZDSYS20210623091809029). \textit{(Corresponding author: Wenjian Luo.)}}
}

\maketitle

\begin{abstract}
Deep Neural Networks (DNNs) are vulnerable to carefully crafted adversarial examples (AEs) that can cause incorrect predictions, posing serious security risks. While knowledge distillation has been explored as a defense, single-model distillation often suffers from limited robustness. In this paper, we propose \textbf{E-SMAD} (Ensemble of Student Models for Adversarial Defense), a defense method that distills a teacher into multiple student models using label-wise temperature scaling. By assigning different temperatures to non-ground-truth classes, each student learns complementary decision boundaries, and their ensemble stabilizes predictions. We evaluate E-SMAD on MNIST, Fashion-MNIST, and CIFAR-10 under FGSM, BIM, PGD, and C\&W attacks. Experimental results demonstrate that E-SMAD reduces attack success rates (ASR) across multiple attacks while maintaining high clean accuracy. Besides, E-SMAD demonstrates strong robustness across different datasets and attack methods compared with competing defense methods. By combining structured knowledge distillation with ensemble inference, E-SMAD provides a simple and effective defense method for security-critical AI applications.

\end{abstract}


%
\IEEEpeerreviewmaketitle


\section{Introduction}
\IEEEPARstart{A}{rtificial} intelligence (AI) systems powered by deep neural networks (DNNs) have achieved remarkable success across a variety of security-critical applications, including autonomous driving, medical diagnostics, and industrial inspection \cite{ref2, ref3, AIH04400, NGUYEN2025102782, 10105503}. Despite these advances, DNNs are highly vulnerable to adversarial examples (AEs), i.e., carefully crafted perturbations that are often imperceptible to humans but can induce high-confidence misclassifications \cite{ref26,9857594,chen2024comprehensive,liu2026llm}. Such vulnerabilities pose serious risks in real-world applications. For instance, minimal perturbations in road sign images may mislead autonomous vehicles \cite{8578273}, and subtle changes in medical images may result in misdiagnosis of critical anomalies \cite{science.aaw4399}. In this paper, we focus on developing defense methods against adversarial examples for DNNs.

The adversarial vulnerability phenomenon was first systematically demonstrated by Szegedy et al. \cite{ref26}, who showed that imperceptible perturbations could consistently fool DNNs across multiple model architectures and datasets. This finding inspired a variety of attack techniques, ranging from single-step gradient-based methods such as the Fast Gradient Sign Method (FGSM) \cite{ref6} to iterative attacks like Projected Gradient Descent (PGD) \cite{ref5}, Basic Iterative Method (BIM) \cite{ref7}, and Carlini \& Wagner (C\&W) attacks \cite{ref18}. These attacks highlight that adversarial vulnerability is a fundamental property of DNNs rather than a flaw of specific models or datasets.

To defend against adversarial attacks, some strategies have been proposed. Adversarial training \cite{ref5} augments the training process with adversarial examples to improve robustness but often vulnerable to unknown adversarial attacks. Preprocessing-based defenses, such as feature squeezing \cite{ref15}, MagNet \cite{ref16}, and PNDetector \cite{LUO2022108383}, attempt to remove or detect perturbations prior to model prediction, yet they are vulnerable to adaptive attacks \cite{tramer2020adaptive}. Most ensemble-based defenses \cite{ref24,ref25,ref35,ref36,ref37,ref41,ref39} leverage complementary errors across multiple models to reduce overall vulnerability, but existing ensembles are often with limited diversity and practical robustness. Knowledge distillation \cite{ref32} transfers knowledge from a high-capacity teacher to a student, providing regularization that can improve robustness, but prior methods typically consider only a single student with uniform temperature scaling \cite{ref4}, leaving the benefits of structured multi-student ensembles largely unexplored.

Despite these advances, existing adversarial defenses exhibit critical limitations. First, ensemble methods often lack efficient strategies to induce complementary decision boundaries, limiting their potential robustness gains. Second, most ensemble methods need to train entire models simultaneously. Third, knowledge distillation has rarely been integrated with ensemble methods in a principled manner, and the use of uniform temperature scaling may fail to fully exploit the benefits of diverse student learning.

To address these challenges, we propose E-SMAD (Ensemble of Student Models for Adversarial Defense), a novel method that integrates knowledge distillation with an ensemble of student models trained by label-wise temperature scaling. Each student is guided by different temperature assignments that create complementary decision boundaries, and the ensemble stabilizes predictions. Unlike prior methods, E-SMAD systematically constructs diverse student ensembles to achieve robust defense across multiple attacks while preserving high clean accuracy. We evaluate E-SMAD on three datasets under four representative adversarial attacks: FGSM, BIM, PGD, and C\&W. Experimental results show that E-SMAD reduces attack success rates while preserving high clean accuracy in most cases, and is competitive with several defense methods, including Adversarial Diversity Promotion (ADP) \cite{ref37}, Joint Gradient Phase and Magnitude Regularization (GPMR) \cite{ref41}, and Diversity Training with a Gradient Alignment Loss (GAL) \cite{ref39}.

The main contributions of this work are summarized as follows.
\begin{itemize}
    \item We propose E-SMAD, an ensemble-based defensive distillation method that trains multiple student models using label-wise temperature scaling to generate complementary decision boundaries, offering a robust defense for security-critical AI applications. Notably, each student model can be trained independently, allowing for feasible ensemble construction under limited computation resources.
    \item We conduct extensive experiments on three datasets and four representative adversarial attacks, demonstrating that the performance of E-SMAD is competitive with typical ensemble defenses.
\end{itemize}

The rest of the paper is organized as follows. Section II reviews related work on adversarial attacks and defenses. Section III details the proposed E-SMAD method. Section IV presents experimental results and analysis. Section V discusses related issues. Section VI concludes the paper and outlines future research directions.

\section{Related Work}
This section reviews related work on adversarial attacks and defenses. We first introduce representative attack methods, followed by a discussion of major defense strategies, including adversarial training, preprocessing-based defenses, ensemble methods, and distillation-based approaches.

\subsection{Adversarial Attacks}
Adversarial attacks have evolved substantially, transitioning from simple gradient-based heuristics to more principled optimization-driven strategies. Szegedy et al. \cite{ref26} were the first to systematically demonstrate that deep neural networks across diverse architectures could be misled by imperceptible input perturbations, establishing adversarial vulnerability as an inherent property of neural networks rather than an isolated artifact. To formalize this phenomenon, they proposed the L-BFGS attack, which casts adversarial example generation as a constrained optimization problem. Although computationally expensive, this formulation laid critical theoretical foundations for subsequent optimization-based adversarial attacks.

The Fast Gradient Sign Method (FGSM) \cite{ref6}, introduced by Goodfellow et al., pioneered gradient-based adversarial attacks by demonstrating that effective perturbations could be generated through a single, computationally efficient update step. While the simplicity of FGSM made it highly accessible and scalable, its one-step formulation inherently limited its attack strength and robustness. To address this limitation, Kurakin et al. \cite{ref7} proposed the Basic Iterative Method (BIM), which extends FGSM by applying it iteratively with small step sizes. This iterative refinement improves the attack success rate by more precisely navigating the loss landscape; however, it comes at the cost of increased computational overhead, while sometimes yielding only marginal gains in attack success rate (ASR) compared to FGSM.

Projected Gradient Descent (PGD) \cite{ref5}, proposed by Madry et al., represents a significant advancement in iterative adversarial attacks. Its key contribution lies in the incorporation of a projection step, which constrains adversarial perturbations to remain within a prescribed $\epsilon$-ball. Moreover, Madry et al. \cite{ref5} theoretically demonstrated that PGD constitutes a universal first-order adversary, providing a principled foundation for adversarial training. Despite its strong empirical and theoretical performance, the iterative nature of PGD incurs substantial computational overhead, which limits its practicality in real-time or resource-constrained attack scenarios.

Carlini and Wagner \cite{ref18} proposed one of the most influential optimization-based adversarial attack frameworks, commonly referred to as the C\&W attack. This method formulates adversarial example generation as a constrained optimization problem that simultaneously maximizes the attack success rate while minimizing the perturbation magnitude. Owing to its strong attack effectiveness, the C\&W attack is capable of circumventing a wide range of defense mechanisms, including defensive distillation, and has therefore become a standard benchmark for evaluating the robustness of defense strategies. Nevertheless, the high computational complexity of the optimization process, together with its reliance on gradient information, significantly limits its applicability in black-box or real-time attack scenarios.

\subsection{Adversarial Defense}

Adversarial defense has been extensively studied over the past decade, leading to a diverse set of strategies that aim to mitigate the vulnerability of deep neural networks to adversarial examples \cite{ref15,ref16,ref17,ref18,ref19}. Despite continuous progress, existing defenses often suffer from fundamental limitations related to computational cost, generalization across attacks, and the trade-off between clean accuracy and robustness. Broadly, prior work on adversarial defense can be categorized into four main directions: adversarial training, preprocessing-based defenses, ensemble-based defenses, and knowledge distillation-based defenses.

\subsubsection{Adversarial Training}

Adversarial training is widely regarded as one of effective defense methods against adversarial examples. Madry et al. \cite{ref5} formulated adversarial training as a min–max optimization problem by incorporating adversarial examples directly into the training process. This method significantly improves robustness against specific attack types used during training.

Despite its effectiveness, adversarial training suffers from several well-recognized limitations \cite{ijcai2021p591}. First, generating adversarial examples during training introduces substantial computational overhead, often increasing training cost significantly compared to standard training. Second, adversarially trained models tend to generalize poorly to unseen attack strategies, limiting their robustness. Third, adversarial training commonly induces a trade-off between clean accuracy and robustness. These limitations hinder the deployment of adversarial training in resource-constrained or accuracy-sensitive applications.

\subsubsection{Preprocessing-based Defenses}

Preprocessing-based defenses aim to mitigate adversarial perturbations by analyzing inputs before classification, without altering the underlying model parameters. Feature squeezing \cite{ref15} reduces the input search space using operations such as bit-depth reduction and spatial smoothing, attempting to expose discrepancies between clean and adversarial samples. While computationally efficient, feature squeezing often suffers from high false-positive rates and could be bypassed by adaptive attackers. MagNet \cite{ref16} adopts a detector–reformer architecture, in which a detector network identifies adversarial inputs and a reformer network attempts to project them back onto the data manifold. Although effective against certain attacks, MagNet becomes less reliable when adversaries jointly optimize against both the detection and reforming mechanisms. Luo et al. \cite{LUO2022108383} proposed PNDetector, an adversarial example detection method based on positive–negative representations, which identifies adversarial examples by comparing the classifier outputs of an input sample and its negative image.

A fundamental limitation of preprocessing-based defenses is their reactive nature. These methods attempt to detect or remove adversarial perturbations rather than improving the intrinsic robustness of the model decision boundaries. As a result, preprocessing-based defenses are particularly vulnerable to adaptive attacks and often fail to provide sustained robustness under strong threat models \cite{tramer2020adaptive}.

\subsubsection{Ensemble-based Defenses}

Ensemble-based defenses exploit the principle that combining multiple models with uncorrelated errors can improve robustness. Perrone and Cooper \cite{ref24} provided a theoretical foundation showing that ensembles reduce overall vulnerability when individual models fail on different subsets of adversarial examples. Empirically, Strauss et al. \cite{ref25} demonstrated that ensemble predictions are more robust than those of single models under adversarial perturbations.  Subsequent studies have explored various ensemble construction strategies. Li et al. \cite{ref35} analyzed the role of error correlation among ensemble members and showed that reduced correlation leads to enhanced robustness. Cheng et al. \cite{ref36} proposed voting-based ensembles trained under diverse settings, while Pang et al. \cite{ref37} introduced Adversarial Diversity Promotion (ADP) to encourage diverse decision boundaries during training. Kariyappa and Qureshi \cite{ref39} proposed Diversity Training with a Gradient Alignment Loss (GAL) regularizer, which explicitly reduces transferability in black-box settings by minimizing the alignment of loss gradients among ensemble members, thereby shrinking the shared adversarial subspace. Dabouei et al. \cite{ref41} further formalized the notion of joint robustness in ensembles and proposed Joint Gradient Phase and Magnitude Regularization (GPMR), which theoretically and empirically improves adversarial robustness by simultaneously promoting gradient direction diversity and equalizing gradient magnitudes among ensemble members, thereby increasing the effective number of classifiers contributing to robustness.

\subsubsection{Distillation-based Defenses}

Knowledge distillation, originally proposed by Hinton et al. \cite{ref32} for model compression, has also been investigated as a defense mechanism \cite{ref4}. By training student models using soft targets from a teacher network, distillation introduces a regularization effect that can smooth decision boundaries.

However, most distillation-based defenses still rely on a single student with uniform temperature scaling, which limits the diversity of learned representations. Although recent works have explored class-aware distillation \cite{ref43} and multi-student settings \cite{ref45}, they do not explicitly leverage temperature design to induce complementary decision boundaries among students for adversarial robustness. As a result, the potential of structured student diversity for ensemble robustness remains largely underexplored.

\section{Method}
\subsection{Threat Model}
\label{subsec:threat_model}

We consider adversarial attacks against multi-class classifiers. Let $f:\mathcal{X}\rightarrow\mathbb{R}^K$ denote a neural network parameterized by $\theta$, which maps an input $x\in\mathcal{X}$ to a vector of logits:
\begin{equation}
    f(x) = \big(f_1(x), \dots, f_K(x)\big),
\end{equation}
where $K$ denotes the number of classes in the classification task.
The logits are transformed into a predictive probability distribution via the softmax function:
\begin{equation}
    p_j(x) = \mathrm{softmax}_j(f(x)) 
    = \frac{\exp(f_j(x))}{\sum_{k=1}^K \exp(f_k(x))},
\end{equation}
where $p_j(x)$ represents the model's predicted probability that input $x$ belongs to class $j$. The final predicted label is
\begin{equation}
\hat{y}(x) = \operatorname*{arg\,max}_{j \in \{1,\dots,K\}} p_j(x).
\end{equation}

We focus on \emph{untargeted adversarial attacks}, in which the adversary does not aim for a specific incorrect class, but only seeks to alter the classifier's prediction such that it differs from the true label \cite{xu2020assessing}. Formally, given a clean input-label pair $(x,y)$, the adversary seeks an adversarial example:
\begin{equation}
    x^{\mathrm{adv}} = x + \delta,
\end{equation}
subject to a perturbation constraint:
\begin{equation}
    \|\delta\|_l \le \epsilon,
\end{equation}
such that the classifier's predicted label is incorrect:
\begin{equation}
    \hat{y}(x^{\mathrm{adv}}) \neq y.
\end{equation}

The goal of the defense is to ensure that the classifier's predicted label remains correct under such perturbations, while maintaining high accuracy on clean inputs. Formally, for a defended model $f_\mathrm{def}$ and any $(x,y)$, we aim to satisfy
\begin{equation}
    \hat{y}_\mathrm{def}(x^{\mathrm{adv}}) = \hat{y}_\mathrm{def}(x) = y, \quad \forall \|\delta\|_l \le \epsilon,
\end{equation}
where $x^{\mathrm{adv}} = x + \delta$ is any adversarial example generated under the attacker's capabilities \cite{ref6,ref5,ref7,ref18}, and $\hat{y}_\mathrm{def}(x)$ is the predicted label of the defended model $f_\mathrm{def}$ on input $x$.

\subsection{Overview of E-SMAD}
\label{subsec:overview}

\begin{figure*}[!t]
\centering
\includegraphics[width=\textwidth]{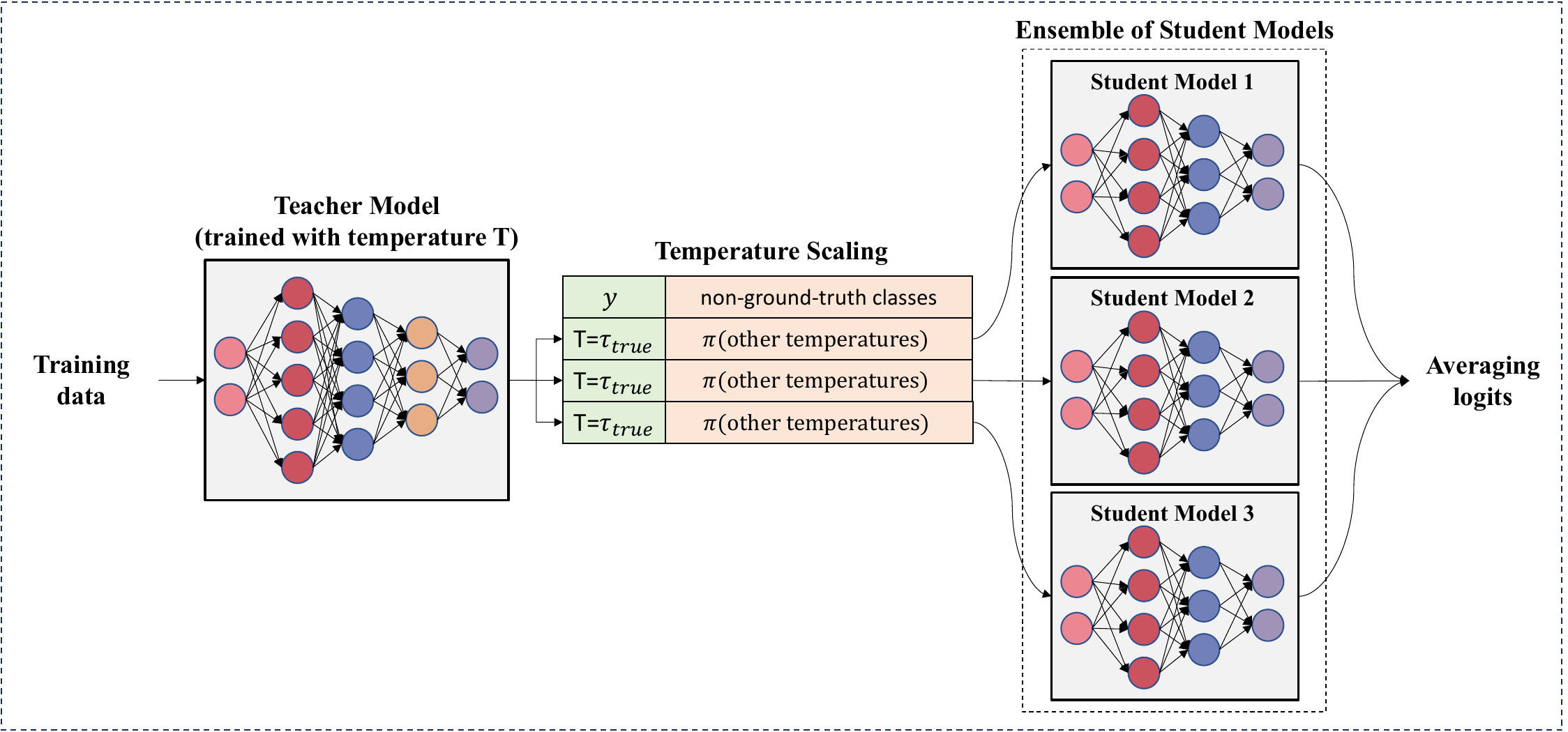}
\caption{Overview of the E-SMAD method. The teacher model distills knowledge to multiple student models using our temperature scaling strategy, which applies different temperatures to different labels.}
\label{fig-1}
\end{figure*}

Fig.~\ref{fig-1} illustrates the overall pipeline of the proposed E-SMAD method. The method follows a teacher--student paradigm combined with ensemble inference to improve adversarial robustness in multi-class classification. A teacher model is first trained on training data to produce reliable logits. Three student models are then trained independently using soft labels derived from the teacher, and the final prediction is obtained by averaging the students' logits, followed by a softmax operation. It is noted that, for better understanding, we just take three student models as an example.

The central design principle of E-SMAD is to introduce structured diversity among student models while preserving strong supervision on the true class. This is achieved through a label-wise temperature scaling strategy, in which different students are encouraged to focus on complementary class-level relationships. By aggregating these complementary students at the logit level, E-SMAD yields a smoother and more robust decision function that is less sensitive to adversarial perturbations. Detailed formulations are provided in the following subsections.

\subsection{Teacher Model Training}
\label{subsec:teacher_training}

The teacher model serves as the source of supervision for all subsequent student models and is therefore required to produce accurate and stable class confidence estimates. The teacher is trained on the training dataset using supervised learning with temperature-scaled softmax outputs.

Formally, given a training dataset $\mathcal{D}=\{(x_i,y_i)\}_{i=1}^{N}$ with $y_i\in\{1,\dots,K\}$, the teacher model is optimized by minimizing the temperature-scaled cross-entropy loss:
\begin{equation}
\mathcal{L}_{\mathrm{teacher}} =
-\frac{1}{N}\sum_{i=1}^{N} \log p_{y_i}(x_i),
\end{equation}
where
\begin{equation}
p_j(x_i)=
\frac{\exp(f_j(x_i)/\tau)}
{\sum_{k=1}^{K}\exp(f_k(x_i)/\tau)}
\end{equation}
denotes the temperature-scaled softmax probability derived from the teacher logits $f(x_i)$, and $\tau$ is the temperature parameter.

The teacher model is trained until convergence and then kept fixed during the subsequent student training stage. By introducing temperature scaling during teacher training, the output probability distribution becomes softer and better captures inter-class similarity information rather than producing overly confident one-hot predictions. These teacher-generated logits are subsequently used to construct label-wise soft labels for training student models.

\subsection{Student Training with Label-wise Temperature Scaling}
\label{subsec:student_training}

We train multiple student models to learn from a common teacher while exhibiting complementary behaviors that improve adversarial robustness when aggregated. To this end, we introduce a label-wise temperature scaling strategy that controls how each student interprets the teacher's confidence on different classes. This subsection describes the temperature scaling rule, the construction of soft targets, and the student optimization objective.

\subsubsection{Label-wise Temperature Scaling}
\label{subsubsec:temperature_assignment}
Label-wise temperature scaling is designed to generate diverse softened supervisory signals by assigning different temperature values to different classes when distilling knowledge from the teacher model. The key idea is to preserve a consistent and confident supervision for the ground-truth class while introducing controlled variability in non-ground-truth class temperatures across different student models.

We present the label-wise temperature scaling strategy as follows.
First, we define a temperature list, where each element specifies the temperature applied to each class when constructing soft targets from the teacher model:
\begin{equation}
    \boldsymbol{\tau} = [\tau_1, \tau_2, \dots, \tau_K].
\end{equation}
For each student model $S^{(m)}$, the smallest temperature $\tau_{\mathrm{true}} = \min(\boldsymbol{\tau})$ is assigned to the ground-truth class to preserve a sharper and more confident supervision signal for the correct label:
\begin{equation}
    \tau^{(m)}_y = \tau_{\mathrm{true}},
\end{equation}
where $\tau^{(m)}_y$ denotes the temperature assigned to the ground-truth class $y$ for student model $S^{(m)}$.

For the remaining classes, we construct a temperature list by removing $\tau_{\mathrm{true}}$ from $\boldsymbol{\tau}$:
\begin{equation}
    \mathcal{T} = \boldsymbol{\tau} \setminus [\tau_{\mathrm{true}}],
\end{equation}
and sample a random permutation of $\mathcal{T}$ for each student:
\begin{equation}
    \pi^{(m)}(\mathcal{T}) = [\tilde{\tau}^{(m)}_1, \dots, \tilde{\tau}^{(m)}_{K-1}].
\end{equation}

Let
\begin{equation}
    \mathcal{C}_y = [1,2,\dots,K] \setminus [y]
\end{equation}
denotes the ordered list of class indices excluding the ground-truth label $y$.
The temperature assignment for non-ground-truth classes is defined as:
\begin{equation}
    \tau^{(m)}_{\mathcal{C}_y[k]} = \tilde{\tau}^{(m)}_k,
    \quad k = 1,2,\dots,K-1.
\end{equation}

This design ensures that the ground-truth class always receives the smallest temperature, while non-ground-truth classes are assigned randomly permuted higher temperatures, leading to diverse softened supervision across different student models.

\subsubsection{Soft Target Construction}
\label{subsubsec:soft_target}

Given the teacher logits $f^T(x)$ and the temperature vector $\boldsymbol{\tau}^{(m)}$, we construct softened teacher logits for student $S^{(m)}$ via element-wise scaling:
\begin{equation}
    \tilde{f}^{T,(m)}_j(x) = \frac{f^T_j(x)}{\tau^{(m)}_j},
    \qquad j = 1, \dots, K.
\end{equation}
The resulting soft target distribution is obtained by applying the softmax function:
\begin{equation}
    q^{(m)}_j(x) =
    \frac{\exp\big(\tilde{f}^{T,(m)}_j(x)\big)}
    {\sum_{k=1}^{K} \exp\big(\tilde{f}^{T,(m)}_k(x)\big)}.
\end{equation}

By assigning larger temperatures to certain non-true classes, the corresponding logits are flattened and contribute less sharply to the soft target distribution, while smaller temperatures preserve finer-grained confidence information. Since each student employs a different temperature permutation, the resulting soft targets encode complementary inter-class relationships across students.

\subsubsection{Student Optimization Objective}
\label{subsubsec:student_objective}

Each student model $S^{(m)}$ is trained using a combination of the standard cross-entropy loss with respect to the true labels and a distillation loss that encourages matching the teacher’s soft target distribution under the label-wise temperature setting. Let $f^{S^{(m)}}(x)$ denote the logits of student $S^{(m)}$ for input $x$, and let $\boldsymbol{\tau}^{(m)}$ be the temperature vector assigned to each class as described in the previous subsection.

We first compute the temperature-scaled student logits:
\begin{equation}
    \tilde{f}^{S^{(m)}}_j(x) = \frac{f^{S^{(m)}}_j(x)}{\tau^{(m)}_j}, \quad j = 1,\dots,K,
\end{equation}
and the corresponding predictive distribution:
\begin{equation}
    p^{(m)}_j(x) =
    \frac{\exp\big(\tilde{f}^{S^{(m)}}_j(x)\big)}
    {\sum_{k=1}^{K} \exp\big(\tilde{f}^{S^{(m)}}_k(x)\big)}.
\end{equation}

The distillation loss for student $S^{(m)}$ is defined as the Kullback--Leibler divergence between the temperature-scaled teacher soft targets $q^{(m)}(x)$ and the student predictions:
\begin{equation}
    \mathcal{L}_{\mathrm{distill}}^{(m)} =
    \frac{1}{N} \sum_{i=1}^{N} 
    \mathrm{KL}\!\left(q^{(m)}(x_i) \;\Vert\; p^{(m)}(x_i)\right).
\end{equation}

The hard-label cross-entropy loss is defined as:
\begin{equation}
    \mathcal{L}_{\mathrm{CE}}^{(m)} =
    - \frac{1}{N} \sum_{i=1}^{N} \log p^{S^{(m)}}_{y_i}(x_i),
\end{equation}
where $p^{S^{(m)}}(x_i)$ denotes the standard softmax of the student logits (without temperature scaling).

The total training loss for each student is a weighted combination of the two terms:
\begin{equation}
    \mathcal{L}^{(m)}_{\mathrm{total}} = 
    \alpha \, \mathcal{L}_{\mathrm{CE}}^{(m)} + 
    (1 - \alpha) \, \mathcal{L}_{\mathrm{distill}}^{(m)},
\end{equation}
where $\alpha \in [0,1]$ controls the balance between supervision from the ground-truth labels and knowledge distilled from the teacher.

This combined objective ensures that each student preserves accuracy on the true class while also capturing the class-level confidence structure encoded in the teacher’s soft targets. By training multiple students with complementary label-wise temperature permutations, the resulting ensemble exhibits diverse error patterns, which is critical for improving robustness when aggregated at inference.

\subsection{Ensemble Inference}
\label{subsec:ensemble_inference}

At test time, we aggregate the predictions of all trained student models to obtain a more robust final output. Let $\{S^{(m)}\}_{m=1}^M$ denote a set of $M$ student models. For an input $x$, each model produces logits $f^{S^{(m)}}(x)$. Rather than averaging the predicted probabilities, we perform logit-level aggregation, which preserves the pre-softmax consensus among students and better maintains inter-class confidence relationships \cite{ref32}.

The aggregated logits are computed as:
\begin{equation}
    \bar{f}(x) = \frac{1}{M} \sum_{m=1}^{M} f^{S^{(m)}}(x),
\end{equation}
and the final predictive probability distribution is obtained via the softmax function:
\begin{equation}
    \bar{p}_j(x) = \frac{\exp(\bar{f}_j(x))}{\sum_{k=1}^{K} \exp(\bar{f}_k(x))}, \qquad j = 1,\dots,K.
\end{equation}
The predicted class is then
\begin{equation}
    \hat{y}(x) = \operatorname*{arg\,max}_{j \in \{1,\dots,K\}} \bar{p}_j(x).
\end{equation}

Logit averaging has two key advantages over probability averaging or majority voting: (1) It preserves the relative magnitudes of student logits, which retains fine-grained class confidence and reduces over-smoothing. (2) It allows the ensemble to exploit the complementary knowledge captured by different students trained under label-wise temperature permutations, resulting in smoother decision boundaries that are empirically less sensitive to small adversarial perturbations.

\section{Experiments}
This section presents a comprehensive evaluation of the E-SMAD method. It first describes the experimental setup, including datasets, model architectures, training details, adversarial attacks, comparative defense methods, and evaluation metrics, and then provides a detailed analysis of experimental results along with comparisons to several defense methods.

\subsection{Experimental Setup}
\subsubsection{Datasets}
\label{subsubsec:datasets}

We evaluate the proposed method on three widely used image classification benchmarks that exhibit increasing levels of visual complexity and input dimensionality. MNIST~\cite{ref47} serves as a low-complexity baseline, consisting of 70,000 grayscale images of handwritten digits (28$\times$28 pixels) from 10 classes, with 60,000 training samples and 10,000 test samples. Fashion-MNIST~\cite{ref52} provides a more challenging grayscale benchmark with the same data split and image resolution as MNIST, but featuring ten categories of fashion products with substantially more complex intra-class variations. CIFAR-10 represents a high-complexity, color image benchmark, containing 60,000 RGB images of size 32$\times$32 from 10 object classes, with 50,000 training images and 10,000 test images.

These datasets are selected to systematically assess the effectiveness and scalability of adversarial defenses across different regimes of visual complexity. MNIST enables analysis under a relatively simple data distribution, Fashion-MNIST evaluates robustness on moderately complex and more realistic visual patterns, while CIFAR-10 poses a significantly harder setting due to its higher dimensionality, richer textures, and greater semantic variability. This progressive evaluation allows us to examine how the proposed defense generalizes from simple to more challenging classification tasks.

\subsubsection{Model Architectures}
The teacher model is implemented as a ResNet-like convolutional neural network tailored for small-resolution image classification, and the student models adopt the same architecture as the teacher. The number of input channels is determined by the dataset, where MNIST and Fashion-MNIST use single-channel grayscale inputs, while CIFAR-10 uses three-channel RGB images. The network begins with a $3\times3$ convolution layer with 64 output channels, stride 1, padding 1, and no bias term, followed by Batch Normalization, ReLU activation, and a Dropout layer with dropout probability $p=0.15$. The backbone then consists of three groups of residual blocks with output channel dimensions of 64, 128, and 256, respectively, where each group contains two residual blocks. After feature extraction, global average pooling is used for feature aggregation, followed by feature flattening, a Dropout layer with $p=0.25$, and a fully connected classification layer. This architecture provides sufficient representational capacity to achieve high accuracy.

\subsubsection{Training Details}
\label{subsec:training_details}
All models are optimized using the Adam optimizer with a learning rate of 0.001, weight decay of 0.0001, and a batch size of 128. Training is performed for up to 200 epochs, with early stopping based on validation accuracy to prevent overfitting. Unless stated otherwise, identical optimization settings are used across all methods to ensure fair and consistent comparisons.

The training procedure follows a three-stage pipeline: (1) teacher model training, (2) independent student model training via knowledge distillation, and (3) ensemble inference. In the first stage, the teacher model is trained on clean data with a temperature of 20. In the second stage, multiple student models are trained independently based on soft targets generated by the fixed teacher model, as detailed in Section~\ref{subsec:student_training}.
Student models are trained under three temperature configurations, corresponding to the ranges $[2,11]$, $[10,19]$, and $[20,29]$. The final ensemble stage does not involve additional training and is applied exclusively at inference time.

\subsubsection{Adversarial Attacks}
\label{subsec:attack_settings}
To evaluate the robustness of the proposed defense, we consider four widely used white-box adversarial attacks: Fast Gradient Sign Method (FGSM) \cite{ref6}, Basic Iterative Method (BIM) \cite{ref7}, Projected Gradient Descent (PGD) \cite{ref5}, and the Carlini \& Wagner (C\&W) attack \cite{ref18}. All attacks are conducted in an untargeted setting. For FGSM, the perturbation budget is set to $\epsilon=0.1$. BIM is performed under the same perturbation constraint $\epsilon=0.1$ with a step size of $\alpha=0.01$ and $10$ iterations. PGD adopts the same perturbation budget and step size, i.e., $\epsilon=0.1$ and $\alpha=0.01$, with $10$ iterations. For the C\&W attack, the initial constant $c$ is set to $1.0$, the number of binary search steps is $9$, and the maximum number of iterations for each optimization round is $100$.

\subsubsection{Comparative Defense Methods}
We conduct a comprehensive comparison of E-SMAD with several representative adversarial ensemble defense methods, including Adversarial Diversity Promotion (ADP) \cite{ref37}, Joint Gradient Phase and Magnitude Regularization (GPMR) \cite{ref41}, and Diversity Training with a Gradient Alignment Loss (GAL) \cite{ref39}.

\subsubsection{Evaluation Metrics}
\label{subsec:evaluation_metrics}

We evaluate the performance of all models using two complementary metrics: clean accuracy (Clean Acc.) and the attack success rate (ASR).

Clean accuracy measures the standard classification performance on unperturbed test examples. It is defined as the fraction of test samples correctly classified:
\begin{equation}
   \mathrm{Clean\;Acc.} = \frac{1}{N} \sum_{i=1}^{N} \mathbf{1}\big(\hat{y}(x_i) = y_i\big),
\end{equation}
where $N$ is the total number of test samples, $x_i$ is an input, $y_i$ is the ground-truth label, and $\hat{y}(x_i)$ is the predicted label.

The attack success rate quantifies the vulnerability of a model to adversarial attacks. It measures the fraction of originally correctly classified test samples that become misclassified after adversarial attacks:
\begin{equation}
\mathrm{ASR}=
\frac{\displaystyle \sum_{i=1}^{N} \mathbf{1}\big(\hat{y}(x_i) = y_i \ \land\ \hat{y}(x_i^{\mathrm{adv}}) \neq y_i\big)}
{\displaystyle\sum_{i=1}^{N} \mathbf{1}\big(\hat{y}(x_i)=y_i\big)},
\end{equation}
where $x_i^{\mathrm{adv}}$ denotes the adversarially perturbed version of $x_i$. A lower ASR indicates stronger robustness to adversarial attacks.

\subsection{Experimental Results}
\subsubsection{Defense Performance of E-SMAD}

Tables~\ref{tab:result1}, \ref{tab:result2}, and \ref{tab:result3} report the defense performance of E-SMAD under different temperature settings across MNIST, Fashion-MNIST, and CIFAR-10. We evaluate both clean accuracy (Clean Acc.) and attack success rate (ASR) under four representative adversarial attacks, i.e., FGSM, BIM, PGD, and C\&W. The results compare three settings: the teacher model, the average of student models, and the proposed E-SMAD. The final row in each table summarizes the number of best results achieved by each method across all attack scenarios.

\begin{table*}[!t]
\centering
\caption{Clean accuracy and attack success rate under temperature range [2, 11]}
\label{tab:result1}
\resizebox{0.9\textwidth}{!}{
\renewcommand{\arraystretch}{1.4}
\begin{tabular}{|c|c|c|c|c|c|c|c|}
\hline
\multicolumn{1}{|c|}{\multirow{2}{*}{\textbf{Dataset}}} & \multirow{2}{*}{\textbf{Attack Method}} &
\multicolumn{2}{c|}{\textbf{Teacher Model}} &
\multicolumn{2}{c|}{\textbf{Average of Student Models}} &
\multicolumn{2}{c|}{\textbf{E-SMAD}} \\
\cline{3-8}
 &  & \textbf{Clean Acc.} & \textbf{ASR} & \textbf{Clean Acc.} & \textbf{ASR} & \textbf{Clean Acc.} & \textbf{ASR} \\
\hline

\multirow{4}{*}{MNIST}
& FGSM & 99.59\% & 7.70\% & 99.67\% & 7.13\% & \textbf{99.73\%} & \textbf{5.10\%} \\ \cline{2-8}
& BIM  & 99.59\% & 22.30\% & 99.67\% & 15.43\% & \textbf{99.73\%} & \textbf{10.20\%} \\ \cline{2-8}
& PGD  & 99.59\% & 54.20\% & 99.67\% & 27.07\% & \textbf{99.73\%} & \textbf{16.00\%} \\ \cline{2-8}
& C\&W & 99.59\% & \textbf{30.50\%} & 99.67\% & 36.03\% & \textbf{99.73\%} & 34.20\% \\
\hline

\multirow{4}{*}{Fashion-MNIST}
& FGSM & \textbf{94.29\%} & 88.00\% & 93.82\% & 74.07\% & 94.27\% & \textbf{73.60\%} \\ \cline{2-8}
& BIM  & \textbf{94.29\%} & 100.00\% & 93.82\% & \textbf{97.47\%} & 94.27\% & 97.50\% \\ \cline{2-8}
& PGD  & \textbf{94.29\%} & 100.00\% & 93.82\% & \textbf{99.20\%} & 94.27\% & 99.70\% \\ \cline{2-8}
& C\&W & \textbf{94.29\%} & 99.50\% & 93.82\% & 99.03\% & 94.27\% & \textbf{98.20\%} \\
\hline

\multirow{4}{*}{CIFAR-10}
& FGSM & 90.29\% & 90.80\% & 89.47\% & \textbf{87.33\%}  & \textbf{91.51\%} & 87.80\% \\ \cline{2-8}
& BIM  & 90.29\% & \textbf{99.50\%} & 89.47\% & 99.57\% & \textbf{91.51\%} & 99.60\% \\ \cline{2-8}
& PGD  & 90.29\% & \textbf{99.50\%} & 89.47\% & 99.97\% & \textbf{91.51\%} & 99.90\% \\ \cline{2-8}
& C\&W & 90.29\% & \textbf{99.50\%} & 89.47\% & 100.00\% & \textbf{91.51\%} & 100.00\% \\
\hline

\multicolumn{2}{|c|}{\textbf{Number of Best Results}} &
4 & 4 & 0 & 3 & \textbf{8} & \textbf{5} \\
\hline
\end{tabular}}
\end{table*}

\begin{table*}[!t]
\centering
\caption{Clean accuracy and attack success rate under temperature range [10, 19]}
\label{tab:result2}
\resizebox{0.9\textwidth}{!}{
\renewcommand{\arraystretch}{1.4}
\begin{tabular}{|c|c|c|c|c|c|c|c|}
\hline
\multicolumn{1}{|c|}{\multirow{2}{*}{\textbf{Dataset}}} & \multirow{2}{*}{\textbf{Attack Method}} &
\multicolumn{2}{c|}{\textbf{Teacher Model}} &
\multicolumn{2}{c|}{\textbf{Average of Student Models}} &
\multicolumn{2}{c|}{\textbf{E-SMAD}} \\
\cline{3-8}
 &  & \textbf{Clean Acc.} & \textbf{ASR} & \textbf{Clean Acc.} & \textbf{ASR} & \textbf{Clean Acc.} & \textbf{ASR} \\
\hline

\multirow{4}{*}{MNIST}
& FGSM & 99.59\% & 7.70\% & 99.68\% & 2.23\% & \textbf{99.73\%} & \textbf{1.80\%} \\ \cline{2-8}
& BIM  & 99.59\% & 22.30\% & 99.68\% & 5.30\% & \textbf{99.73\%} & \textbf{3.70\%} \\ \cline{2-8}
& PGD  & 99.59\% & 54.20\% & 99.68\% & 6.87\% & \textbf{99.73\%} & \textbf{4.10\%} \\ \cline{2-8}
& C\&W & 99.59\% & \textbf{30.50\%} & 99.68\% & 33.77\% & \textbf{99.73\%} & 32.80\% \\
\hline

\multirow{4}{*}{Fashion-MNIST}
& FGSM & 94.29\% & 88.00\% & 94.04\% & 45.77\% & \textbf{94.34\%} & \textbf{43.40\%} \\ \cline{2-8}
& BIM  & 94.29\% & 100.00\% & 94.04\% & 67.27\% & \textbf{94.34\%} & \textbf{67.00\%} \\ \cline{2-8}
& PGD  & 94.29\% & 100.00\% & 94.04\% & 94.60\% & \textbf{94.34\%} & \textbf{93.90\%} \\ \cline{2-8}
& C\&W & 94.29\% & 99.50\% & 94.04\% & 98.50\% & \textbf{94.34\%} & \textbf{97.00\%} \\
\hline

\multirow{4}{*}{CIFAR-10}
& FGSM & 90.29\% & 90.80\% & 89.14\% & 64.73\% & \textbf{91.00\%} & \textbf{63.30\%} \\ \cline{2-8}
& BIM  & 90.29\% & 99.50\% & 89.14\% & \textbf{69.10\%} & \textbf{91.00\%} & 69.90\% \\ \cline{2-8}
& PGD  & 90.29\% & 99.50\% & 89.14\% & \textbf{94.60\%} & \textbf{91.00\%} & 94.90\% \\ \cline{2-8}
& C\&W & 90.29\% & \textbf{99.50\%} & 89.14\% & 100.00\% & \textbf{91.00\%} & 100.00\% \\
\hline

\multicolumn{2}{|c|}{\textbf{Number of Best Results}} &
0 & 2 & 0 & 2 & \textbf{12} & \textbf{8} \\
\hline
\end{tabular}}
\end{table*}

\begin{table*}[!t]
\centering
\caption{Clean accuracy and attack success rate under temperature range [20, 29]}
\label{tab:result3}
\resizebox{0.9\textwidth}{!}{
\renewcommand{\arraystretch}{1.4}
\begin{tabular}{|c|c|c|c|c|c|c|c|}
\hline
\multicolumn{1}{|c|}{\multirow{2}{*}{\textbf{Dataset}}} & \multirow{2}{*}{\textbf{Attack Method}} &
\multicolumn{2}{c|}{\textbf{Teacher Model}} &
\multicolumn{2}{c|}{\textbf{Average of Student Models}} &
\multicolumn{2}{c|}{\textbf{E-SMAD}} \\
\cline{3-8}
 &  & \textbf{Clean Acc.} & \textbf{ASR} & \textbf{Clean Acc.} & \textbf{ASR} & \textbf{Clean Acc.} & \textbf{ASR} \\
\hline

\multirow{4}{*}{MNIST}
& FGSM & 99.59\% & 7.70\% & \textbf{99.67\%} & 1.40\% & \textbf{99.67\%} & \textbf{1.10\%} \\ \cline{2-8}
& BIM  & 99.59\% & 22.30\% & \textbf{99.67\%} & 2.17\% & \textbf{99.67\%} & \textbf{1.80\%} \\ \cline{2-8}
& PGD  & 99.59\% & 54.20\% & \textbf{99.67\%} & 2.60\% & \textbf{99.67\%} & \textbf{1.90\%} \\ \cline{2-8}
& C\&W & 99.59\% & \textbf{30.50\%} & \textbf{99.67\%} & \textbf{30.50\%} & \textbf{99.67\%} & 30.90\% \\
\hline

\multirow{4}{*}{Fashion-MNIST}
& FGSM & 94.29\% & 88.00\% & 94.43\% & 24.27\% & \textbf{94.70\%} & \textbf{22.90\%} \\ \cline{2-8}
& BIM  & 94.29\% & 100.00\% & 94.43\% & 32.30\% & \textbf{94.70\%} & \textbf{32.00\%} \\ \cline{2-8}
& PGD  & 94.29\% & 100.00\% & 94.43\% & 79.70\% & \textbf{94.70\%} & \textbf{78.90\%} \\ \cline{2-8}
& C\&W & 94.29\% & 99.50\% & 94.43\% & 96.77\% & \textbf{94.70\%} & \textbf{94.30\%} \\
\hline

\multirow{4}{*}{CIFAR-10}
& FGSM & 90.29\% & 90.80\% & 90.65\% & 36.43\% & \textbf{91.83\%} & \textbf{34.90\%} \\ \cline{2-8}
& BIM  & 90.29\% & 99.50\% & 90.65\% & 41.83\% & \textbf{91.83\%} & \textbf{40.50\%} \\ \cline{2-8}
& PGD  & 90.29\% & 99.50\% & 90.65\% & \textbf{87.80\%} & \textbf{91.83\%} & 88.70\% \\ \cline{2-8}
& C\&W & 90.29\% & \textbf{99.50\%} & 90.65\% & 100.00\% & \textbf{91.83\%} & 100.00\% \\
\hline

\multicolumn{2}{|c|}{\textbf{Number of Best Results}} &
0 & 2 & 4 & 2 & \textbf{12} & \textbf{9} \\
\hline
\end{tabular}}
\end{table*}

Overall, E-SMAD consistently achieves the best or near-best performance in terms of clean accuracy while substantially reducing ASR across most attack scenarios. Compared with both the teacher and the averaged student baselines, E-SMAD demonstrates a more favorable robustness-accuracy trade-off, indicating that the proposed distillation-based ensemble effectively aggregates complementary robustness information from multiple student models.

\textbf{Dataset-wise analysis.} On MNIST, all methods achieve nearly saturated clean accuracy (near 100\%), so robustness becomes the primary differentiating factor. E-SMAD consistently achieves the lowest ASR under FGSM, BIM, and PGD attacks across all temperature settings. Under temperature range [10, 19], the PGD ASR decreases dramatically from 54.20\% for the teacher model to only 4.10\%, while BIM ASR drops from 22.30\% to 3.70\%. Under temperature range [20, 29], the robustness improvement becomes more pronounced, with PGD ASR reduced further to 1.90\%. These results indicate that E-SMAD is highly effective at enhancing robustness on low-complexity datasets, even when clean accuracy is already saturated. However, under the optimization-based C\&W attack, all methods exhibit relatively higher ASR, implying that stronger optimization attacks remain challenging even for ensemble-based defenses.

On Fashion-MNIST, which contains more ambiguous class boundaries, the advantage of E-SMAD becomes more evident as the temperature increases. Under temperature range [2, 11], robustness improvements are relatively limited, especially under PGD and C\&W where ASR remains close to 100\%. In contrast, under temperature ranges [10, 19] and [20, 29], E-SMAD substantially reduces ASR while simultaneously improving clean accuracy. For example, under FGSM, ASR decreases from 88.00\% (teacher) to 43.40\% under [10, 19] and further to 22.90\% under [20, 29]. Similarly, BIM ASR decreases from 100.00\% to 32.00\% under [20, 29]. These results suggest that high-temperature distillation significantly smooths decision boundaries and improves robustness generalization. Nevertheless, under stronger iterative attacks such as PGD and C\&W, the ASR remains relatively high, indicating that these attacks continue to pose significant challenges for robustness improvement on datasets with ambiguous decision boundaries.

On CIFAR-10, which is substantially more complex due to higher dimensionality and richer semantic variability, E-SMAD consistently improves clean accuracy while also achieving higher adversarial robustness than the teacher model. Unlike MNIST and Fashion-MNIST, clean accuracy gains on CIFAR-10 are more noticeable, increasing from 90.29\% (teacher) to as high as 91.83\% under temperature range [20, 29]. Meanwhile, robustness improvements become progressively stronger as the temperature increases. Under FGSM, ASR decreases from 90.80\% to 63.30\% under temperature range [10, 19] and further to 34.90\% under temperature range [20, 29]. Similarly, BIM ASR is reduced from 99.50\% to 40.50\% under temperature range [20, 29]. These results demonstrate that E-SMAD benefits significantly from diversified temperature distillation on high-complexity datasets. However, under strong attacks such as PGD and C\&W, ASR remains high across all methods, especially for C\&W where ASR approaches 100\%. This suggests that although E-SMAD improves robustness substantially, defending against highly optimized iterative attacks on complex datasets remains an open challenge.

\textbf{Effect of temperature scaling.} The experimental results reveal a clear relationship between temperature range and robustness performance. Compared with lower temperatures ([2, 11]), higher temperature ranges ([10, 19] and especially [20, 29]) consistently produce lower ASR across nearly all datasets and attacks. Higher temperatures soften the teacher logits and encourage student models to learn richer inter-class relationships, thereby improving robustness transfer. Moreover, E-SMAD maintains or even improves clean accuracy as the temperature increases, indicating that robustness gains are not obtained at the expense of standard classification performance. However, smooth distributions still show limited effectiveness against optimization-intensive attacks such as C\&W, suggesting that temperature scaling alone cannot fully eliminate adversarial vulnerability.

\textbf{Overall conclusion.} Across most attack scenarios, E-SMAD achieves better robustness than the teacher model and generally performs competitively or better than averaged student baselines while maintaining high clean accuracy. The improvements become increasingly pronounced under higher temperatures. E-SMAD is effective in improving the balance between clean accuracy and adversarial robustness across datasets with different levels of complexity.

\subsection{Comparison with Existing Defense Methods}

Table~\ref{tab:comparison} compares E-SMAD with several representative defense methods, including ADP, GPMR, and GAL, across MNIST, Fashion-MNIST, and CIFAR-10 under four adversarial attacks (FGSM, BIM, PGD, and C\&W). We report both clean accuracy and attack success rate (ASR) to evaluate the trade-off between natural performance and adversarial robustness. The temperature range for E-SMAD in this comparison is [20, 29], which provides a good balance based on our previous analysis.

\begin{table*}[!t]
\centering
\caption{Performance Comparison of E-SMAD and Existing Defense Methods}
\label{tab:comparison}
\resizebox{1\textwidth}{!}{%
\renewcommand{\arraystretch}{1.5}
\begin{tabular}{|c|c|c|c|c|c|c|c|c|c|}
\hline
\multicolumn{1}{|c|}{\multirow{2}{*}{\textbf{Dataset}}} & \multirow{2}{*}{\textbf{Attack Method}} & \multicolumn{2}{c|}{\textbf{ADP}} & \multicolumn{2}{c|}{\textbf{GPMR}} & \multicolumn{2}{c|}{\textbf{GAL}} & \multicolumn{2}{c|}{\textbf{E-SMAD}} \\ \cline{3-10} 
\multicolumn{1}{|c|}{} &  & \multicolumn{1}{c|}{\textbf{Clean Acc.}} & \textbf{ASR} & \multicolumn{1}{c|}{\textbf{Clean Acc.}} & \textbf{ASR} & \multicolumn{1}{c|}{\textbf{Clean Acc.}} & \textbf{ASR} & \multicolumn{1}{c|}{\textbf{Clean Acc.}} & \textbf{ASR} \\ \hline

\multicolumn{1}{|c|}{\multirow{4}{*}{\textbf{MNIST}}}
& FGSM & \multicolumn{1}{c|}{\textbf{99.74\%}} & 3.40\% & \multicolumn{1}{c|}{99.59\%} & 3.30\% & \multicolumn{1}{c|}{99.73\%} & 46.30\% & \multicolumn{1}{c|}{99.67\%} & \textbf{1.10\%} \\ \cline{2-10}

\multicolumn{1}{|c|}{}
& BIM & \multicolumn{1}{c|}{\textbf{99.74\%}} & 6.30\% & \multicolumn{1}{c|}{99.59\%} & 4.50\% & \multicolumn{1}{c|}{99.73\%} & 58.70\% & \multicolumn{1}{c|}{99.67\%} & \textbf{1.80\%} \\ \cline{2-10}

\multicolumn{1}{|c|}{}
& PGD & \multicolumn{1}{c|}{\textbf{99.74\%}} & 8.70\% & \multicolumn{1}{c|}{99.59\%} & 3.70\% & \multicolumn{1}{c|}{99.73\%} & 70.40\% & \multicolumn{1}{c|}{99.67\%} & \textbf{1.90\%} \\ \cline{2-10}

\multicolumn{1}{|c|}{}
& C\&W & \multicolumn{1}{c|}{\textbf{99.74\%}} & \textbf{16.50\%} & \multicolumn{1}{c|}{99.59\%} & 21.30\% & \multicolumn{1}{c|}{99.73\%} & 20.90\% & \multicolumn{1}{c|}{99.67\%} & 30.90\% \\ \hline

\multicolumn{1}{|c|}{\multirow{4}{*}{\textbf{Fashion-MNIST}}}
& FGSM & \multicolumn{1}{c|}{94.64\%} & 80.70\% & \multicolumn{1}{c|}{90.62\%} & 48.90\% & \multicolumn{1}{c|}{92.24\%} & 47.90\% & \multicolumn{1}{c|}{\textbf{94.70\%}} & \textbf{22.90\%} \\ \cline{2-10}

\multicolumn{1}{|c|}{}
& BIM & \multicolumn{1}{c|}{94.64\%} & 100.00\% & \multicolumn{1}{c|}{90.62\%} & 60.40\% & \multicolumn{1}{c|}{92.24\%} & 71.90\% & \multicolumn{1}{c|}{\textbf{94.70\%}} & \textbf{32.00\%} \\ \cline{2-10}

\multicolumn{1}{|c|}{}
& PGD & \multicolumn{1}{c|}{94.64\%} & 100.00\% & \multicolumn{1}{c|}{90.62\%} & \textbf{54.20\%} & \multicolumn{1}{c|}{92.24\%} & 72.90\% & \multicolumn{1}{c|}{\textbf{94.70\%}} & 78.90\% \\ \cline{2-10}

\multicolumn{1}{|c|}{}
& C\&W & \multicolumn{1}{c|}{94.64\%} & 88.60\% & \multicolumn{1}{c|}{90.62\%} & 89.80\% & \multicolumn{1}{c|}{92.24\%} & \textbf{81.40\%} & \multicolumn{1}{c|}{\textbf{94.70\%}} & 94.30\% \\ \hline

\multicolumn{1}{|c|}{\multirow{4}{*}{\textbf{CIFAR-10}}}
& FGSM & \multicolumn{1}{c|}{\textbf{92.51\%}} & 91.20\% & \multicolumn{1}{c|}{78.87\%} & 97.90\% & \multicolumn{1}{c|}{90.97\%} & 88.00\% & \multicolumn{1}{c|}{91.83\%} & \textbf{34.90\%} \\ \cline{2-10}

\multicolumn{1}{|c|}{}
& BIM & \multicolumn{1}{c|}{\textbf{92.51\%}} & 100.00\% & \multicolumn{1}{c|}{78.87\%} & 100.00\% & \multicolumn{1}{c|}{90.97\%} & 100.00\% & \multicolumn{1}{c|}{91.83\%} & \textbf{40.50\%} \\ \cline{2-10}

\multicolumn{1}{|c|}{}
& PGD & \multicolumn{1}{c|}{\textbf{92.51\%}} & 100.00\% & \multicolumn{1}{c|}{78.87\%} & 100.00\% & \multicolumn{1}{c|}{90.97\%} & 100.00\% & \multicolumn{1}{c|}{91.83\%} & \textbf{88.70\%} \\ \cline{2-10}

\multicolumn{1}{|c|}{}
& C\&W & \multicolumn{1}{c|}{\textbf{92.51\%}} & 100.00\% & \multicolumn{1}{c|}{78.87\%} & 100.00\% & \multicolumn{1}{c|}{90.97\%} & \textbf{96.60\%} & \multicolumn{1}{c|}{91.83\%} & 100.00\% \\ \hline

\multicolumn{2}{|c|}{\textbf{Number of Best Results}}
& 8 & 1
& 0 & 1
& 0 & 2
& \textbf{4} & \textbf{8} \\ \hline

\end{tabular}
}
\end{table*}

Overall, E-SMAD demonstrates the strongest overall robustness among the compared defense methods, achieving the lowest ASR under most attack scenarios while maintaining competitive clean accuracy, which is close to the highest-performing baseline. These results indicate that E-SMAD achieves a superior balance between classification performance and adversarial robustness across different datasets and attack methods.

\textbf{Method-wise comparison.} ADP achieves the highest clean accuracy on MNIST and CIFAR-10, but its robustness is highly unstable under stronger attacks. On Fashion-MNIST and CIFAR-10, the ASR of ADP frequently reaches extremely high values, in some cases rising to 100\%, indicating weak resistance against adversarial attacks despite strong clean performance. GPMR shows relatively good robustness on simpler datasets such as MNIST. However, its clean accuracy decreases significantly on CIFAR-10, and its robustness deteriorates substantially under stronger attacks, suggesting limited scalability and generalization to more complex datasets. GAL exhibits highly inconsistent behavior across datasets and attack methods. Although it achieves moderate robustness under certain settings, it suffers severe robustness degradation on MNIST, where ASRs increase dramatically under BIM and PGD, indicating poor stability across different attack scenarios.

In contrast, E-SMAD maintains strong robustness while preserving high clean accuracy under most attack scenarios. On MNIST, E-SMAD achieves the lowest ASR under FGSM, BIM, and PGD attacks (1.10\%, 1.80\%, and 1.90\%, respectively), significantly outperforming all baseline methods. Although its robustness under the C\&W attack is weaker than ADP, GPMR and GAL, E-SMAD still maintains competitive clean accuracy close to the best baseline. On Fashion-MNIST, E-SMAD achieves the highest clean accuracy (94.70\%) while substantially reducing ASR under FGSM and BIM compared with competing methods, demonstrating a more favorable balance between robustness and accuracy. On CIFAR-10, E-SMAD shows clear robustness advantages under FGSM and BIM attacks, reducing ASR to 34.90\% and 40.50\%, respectively, while preserving high clean accuracy (91.83\%). Although robustness under PGD and C\&W attacks remains limited on this more challenging dataset, E-SMAD still achieves better robustness-accuracy trade-offs than most existing methods.

\textbf{Overall conclusion.} Across all evaluated datasets and attack methods, E-SMAD achieves the largest number of best robustness results while maintaining competitive clean accuracy. The experimental results demonstrate that E-SMAD provides a more stable and effective defense mechanism than existing methods.

\section{Discussions}
\subsection{Effect of Student Model Number}

In the current study, we evaluate E-SMAD using an ensemble of three student models. While this configuration demonstrates strong robustness and high clean accuracy across datasets and attack types, the effect of varying the number of students remains an open question. Intuitively, increasing the ensemble size may further improve robustness by introducing more diverse decision boundaries, whereas reducing the number of students could lower computational cost at the potential expense of adversarial protection. 

A systematic investigation of how student model quantity influences adversarial performance requires additional experimentation. We leave this exploration to future work, as it involves tuning the trade-off between ensemble size, model diversity, and inference overhead. Such studies would provide further insight into the scalability and practical deployment of E-SMAD in resource-constrained settings.

\subsection{Impact of Input Complexity}

The performance of E-SMAD varies across inputs with differing complexity, highlighting the role of input characteristics in evaluating adversarial defenses. On MNIST, which consists of low-dimensional grayscale digits, all student models and the ensemble achieve high clean accuracy and low attack success rate (ASR), reflecting the relative simplicity of the task. Fashion-MNIST introduces more complex intra-class variability, and while individual students exhibit varied robustness, the ensemble effectively stabilizes predictions and reduces ASR. CIFAR-10, with higher-dimensional RGB images and richer semantic content, represents the most challenging setting; although ASR values are higher for all models, the ensemble outperforms student models in most cases, demonstrating the scalability of E-SMAD to more complex inputs.

These observations indicate that input dimensionality and intra-class diversity directly influence adversarial robustness. Evaluating E-SMAD on larger and more diverse datasets, such as ImageNet, would provide further insight into the method’s generalization capabilities and practical applicability.

\subsection{Impact of Model Design}

Input complexity also has important implications for model architecture. In the current study, all student models share the same model architecture, which is sufficient for low-dimensional inputs but may limit diversity in decision boundaries. As input complexity increases, models with deeper or wider layers may be required to adequately capture the underlying feature space and maintain performance. 

Future work could explore heterogeneous student architectures—varying depth, width, or residual connectivity—to enhance ensemble diversity and further improve robustness. Systematic evaluation of different architectures would clarify how model design interacts with distillation-based ensemble, enabling more effective and scalable deployment of E-SMAD across a range of tasks.

\subsection{Computational Efficiency}

The proposed E-SMAD method enhances robustness through an ensemble of student models, which raises concerns regarding computational cost. In our current setting, each student shares the same architecture and capacity as the teacher model, leading to comparable per-model training and inference costs.

During inference, the ensemble requires forward passes through multiple student models. However, these computations can be efficiently parallelized across available hardware resources, which significantly alleviates the overall latency. In practice, the inference time is primarily determined by the slowest model in the parallel execution rather than scaling linearly with the number of students.

Moreover, modern parallel computing infrastructures, such as multi-GPU or distributed systems, make it feasible to deploy such ensembles with manageable overhead. Empirical results indicate that the additional computational cost introduced by E-SMAD remains acceptable, enabling its application in real-time or large-scale scenarios.

\section{Conclusions}
In this work, we proposed E-SMAD, an ensemble-based adversarial defense method that leverages multiple student models trained with label-wise temperature diversification. By distilling knowledge from a teacher into complementary student models and aggregating their predictions at the logit level, E-SMAD effectively stabilizes predictions under a range of adversarial perturbations. Extensive experiments on MNIST, Fashion-MNIST, and CIFAR-10 demonstrate that E-SMAD consistently achieves higher clean accuracy and lower attack success rates. The results highlight the benefits of combining temperature-diversified student models to improve robustness. 

While the current study focuses on a fixed number of students with the same model architecture, future work will explore the impact of heterogeneous student architectures, varying ensemble sizes, and evaluations on more complex datasets.

\bibliographystyle{IEEEtran}
\bibliography{mybibfile.bib}

\end{document}